\documentclass[a4paper]{cas-dc}

\usepackage[numbers]{natbib}
\usepackage{graphicx}
\usepackage{subfigure}
\usepackage{multirow}
\usepackage{color}
\usepackage{times}
\usepackage{soul}
\usepackage{url}
\usepackage{amsthm}
\usepackage{booktabs}
\usepackage{algorithm}
\usepackage{algorithmic}
\usepackage{threeparttable}
\usepackage[justification=centering]{caption}
\usepackage{ulem}
\usepackage{hyperref}
\hypersetup{
	colorlinks=true,
	linkcolor=blue,
	filecolor=blue,      
	urlcolor=blue,
	citecolor=cyan,
}

\usepackage{array}
\usepackage{colortbl}

{}
{}
{}

\def\tsc#1{\csdef{#1}{\textsc{\lowercase{#1}}\xspace}}
\tsc{WGM}
\tsc{QE}
\tsc{EP}
\tsc{PMS}
\tsc{BEC}
\tsc{DE}

\begin{document}
\let\WriteBookmarks\relax
\def\floatpagepagefraction{1}
\def\textpagefraction{.001}
\shorttitle{Learning to Disentangle Scenes for Person Re-identification}
\shortauthors{Zang et~al.}

\title [mode = title]{Learning to Disentangle Scenes for Person Re-identification}                

\tnotemark[1]

\tnotetext[1]{This work was supported by the Key-Area Research and Development Program of Guangdong Province (2019B121204008), the National Natural Science Foundation of China (61801303 and 62031013), the Guangdong Basic and Applied Basic Research Foundation (2019A1515012031), the Shenzhen Science and Technology Plan Basic Research Project (JCYJ20190808161805519), and the Shenzhen Fundamental Research Program (GXWD20201231165807007-20200806163656003).}

\author[1]{Xianghao Zang}[]
\ead{zangxh@pku.edu.cn}
\author[1]{Ge Li}
\ead{geli@ece.pku.edu.cn}
\author[1]{Wei Gao}
\ead{gaowei262@pku.edu.cn}
\author[2]{Xiujun Shu}
\ead{shuxj@pcl.ac.cn}



\address[1]{School of Electronic and Computer Engineering, Peking University, Shenzhen 518055, China.}
\address[2]{Peng Cheng Laboratory, Shenzhen 518034, China.}
\cortext[cor1]{Corresponding author: Wei Gao.}

\begin{abstract}
There are many challenging problems in the person re-identification (ReID) task, such as the occlusion and scale variation. Existing works usually tried to solve them by employing a one-branch network. This one-branch network needs to be robust to various challenging problems, which makes this network overburdened. This paper proposes to divide-and-conquer the ReID task. For this purpose, we employ several self-supervision operations to simulate different challenging problems and handle each challenging problem using different networks. Concretely, we use the random erasing operation and propose a novel random scaling operation to generate new images with controllable characteristics. A general multi-branch network, including one master branch and two servant branches, is introduced to handle different scenes. These branches learn collaboratively and achieve different perceptive abilities. In this way, the complex scenes in the ReID task are effectively disentangled, and the burden of each branch is relieved. The results from extensive experiments demonstrate that the proposed method achieves state-of-the-art performances on three ReID benchmarks and two occluded ReID benchmarks. Ablation study also shows that the proposed scheme and operations significantly improve the performance in various scenes. The code is available at \href{https://git.openi.org.cn/zangxh/LDS.git}{https://git.openi.org.cn/zangxh/LDS.git}.

\end{abstract}

\begin{keywords}
person re-identification \sep divide-and-conquer \sep multi-branch network
\end{keywords}

\maketitle

\section{Introduction}
Person re-identification (ReID) has drawn increasing attention in computer vision society. Given a person image from the query, ReID aims to find all images of the same person from the gallery. In practice, the ReID task has wide- spread applications in social security and surveillance systems. For example, with the help of surveillance cameras, it can help find out the suspect criminals, look for a lost child in a large mall, etc \cite{RN123}.

Despite achieving much progress \cite{RN525} \cite{RN526}, it is still challenging to handle various complex scenes in the ReID task, such as scale variation, occlusion, false detection, and a similar appearance. Most existing approaches can be categorized as the one-branch network. And the challenging problems overburden these one-branch networks. To achieve good performance, these one-branch networks have to utilize sophisticated designs or employ additional information (semantics, pose information, \textit{etc.} \cite{RN345} \cite{RN291} \cite{RN467} \cite{RN229}). These elaborate designs make the one-branch network complicated and over-engineered. 

Recently, multi-branch networks have shown their potentials in many fields of deep learning, such as image classification \cite{RN412} \cite{RN187}, knowledge distillation \cite{RN444} \cite{RN438}, cross-domain/modality learning \cite{RN339} \cite{RN535}. For the ReID task, many multi-branch networks were also proposed. These networks fall into two categories. The first category borrows thoughts from knowledge distillation \cite{RN415} \cite{RN427}. They usually have two networks, teacher and student, and use a two-stage training process. The teacher is trained for complex tasks in the first stage. In the second stage, the teacher network is fixed, and its knowledge is transferred to the student. Although this method has two branches, only one branch learns in each training stage. Therefore, it is still under the paradigm of a one-branch network. The second category employs two equal networks and let them co-teach \cite{RN445} \cite{RN428} \cite{RN339}. Their branches share the same responsibilities and support each other, which produces better performance than their one-branch counterpart. However, each branch still needs to deal with various challenging scenes, and its responsibility has not reduced, resulting in limited performance improvement. 

\begin{figure}[t]
	\centering
	\subfigure[Occluded Scenes]{
		\begin{minipage}{0.23\textwidth} \label{occlusion}
			\centering
			\includegraphics[scale=0.65]{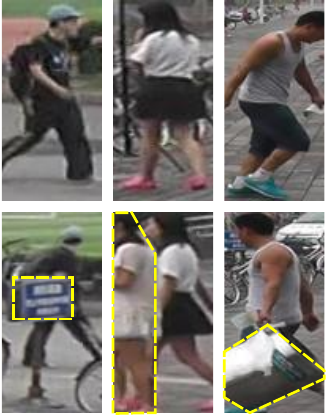}
	\end{minipage}}
	\subfigure[Scale Variation Scenes]{
		\begin{minipage}{0.23\textwidth} \label{scale} 
			\centering
			\includegraphics[scale=0.65]{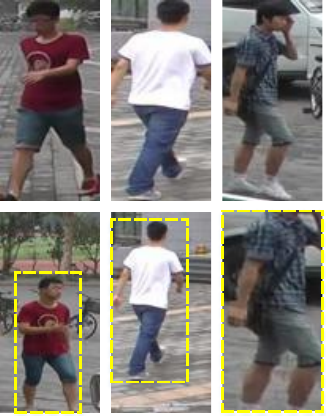}
	\end{minipage}}
	\caption{Two typical challenging scenes in ReID. The top and bottom rows show normal and difficult scenes. (a) Occluded scenes. The occluded areas are within the yellow dotted lines. (b) Scale variation scenes. The sizes of pedestrians in the yellow boxes change.} \label{motivation}
\end{figure}

To effectively disentangle the complex scenes, we propose a \textbf{divide-and-conquer} strategy for the ReID task. We mainly analyze two challenging scenes, \textit{i.e.,} occlusion and scale variation, as illustrated in Fig. \ref{motivation}. We conquer them one by one and improve the overall performance for the ReID task. To this end, we apply two self-supervision operations to the input image to obtain new images with the characteristics of challenging scenes. Concretely, we employ the \textit{random erasing} to generate the occluded scenes and propose \textit{random scaling} to generate the scale variation scenes. In this way, the ReID task is divided into simpler ones. The original image is also kept as the general scenes to provide the missing information for other generated images. We also introduce a new input manner, \textit{i.e.,} \textit{homologous input}. This manner solves the input image misalignment problem and further improves the performance. To conquer each challenging scene, we propose a multi-branch network, as illustrated in Fig. \ref{framework}. There are one master branch and two servant branches in this framework. Each servant branch is assigned to deal with one specific challenge. And the master branch is designed to handle the general scenes. The traditional one-branch network needs to deal with occlusion, scale variation, \textit{etc.} In our framework, each servant branch only needs to deal with occlusion or scale variation. Therefore, the burden of each servant branch is relieved. To train the multi-branch network, we employ mutual learning to transfer the knowledge between different branches. These branches learn collaboratively and promote each other. For the master branch, we use the original input image without any artificial change. The knowledge from the servant branches benefits the master branch and decreases its overfitting possibility for general scenes. For the servant branch, the artificial image loses some information due to the self-supervision operations. The knowledge from the master branch makes the servant branch implicitly learn the missing information and obtain robustness for a specific scene. In the testing process, features from multiple branches are concatenated as the whole feature representation. Since each branch has a different perceptive ability, the concatenated feature is robust for various scenes.

We evaluate the proposed scheme on three ReID benchmarks, including Market1501 \cite{RN149}, DukeMTMC-reID \cite{RN170}, MSMT17 \cite{RN172}, and two large-scale occluded ReID benchmark, P-DukeMTMC-reID \cite{RN475}, and Occluded-DukeMTMC \cite{RN291}. The experiments demonstrate our method achieves state-of-the-art performances. An extensive ablation study also shows that the proposed scheme improves the robustness in various scenes. 

The main contributions of this paper can be summarized as follows:
\begin{itemize}
	\item For the challenging ReID task, we introduce a divide-and-conquer strategy to deal with it, which effectively reduces the network learning burden.
	\item Unlike the traditional multi-branch networks, knowledge communication from different scenes improves the overall performance, which capitalizes on the potentials of multi-branch networks. 
	\item Experiments on three ReID benchmarks and two occluded ReID benchmarks show that our scheme achieves state-of-the-art performances. The ablation study also demonstrates the effectiveness of the proposed scheme in various scenes.
\end{itemize}

The rest of this paper is organized as follows. The related works are reviewed and discussed in Section \ref{related work}, and then we elaborate on the proposed method in Section \ref{method}. Experimental results and analysis are presented in Section \ref{Experiments}, and finally, Section \ref{conlusion} concludes this paper.

\section{Related Work} \label{related work}

\subsection{Person Re-identification}
For the ReID task, the problem of misalignment introduced by scale variation and occlusion has aroused great interest in the computer vision community. Many works explored this problem \cite{RN456} \cite{RN350} \cite{RN178} \cite{RN263} \cite{RN477}. Luo \textit{et.al} \cite{RN263} proposed a Dynamically Matching Local Information (DMLI) to align the local information dynamically. The DMLI calculates the distances between different parts of possible image pairs through a dynamic programming strategy. Miao \textit{et.al} \cite{RN291} employed the pose estimator to generate landmarks. These landmarks are utilized to indicate the model to focus on the non-occluded regions to overcome the noise introduced by the various obstacles. In this way, they obtained aligned feature representations for the ReID task.

The methods above explicitly achieved the feature alignment with the help of various supporting information. Others deal with this problem in a implicit manner. Zhou \textit{et.al} \cite{RN479} proposed Omni-Scale Networks (OSNet), which introduced an aggregating gate to fuse feature from different scales to achieve an omni-scale feature representation. The OSNet handles the misalignment by aggregating the multi-scale features and achieves good performance. Jin \textit{et.al} \cite{RN345} proposed Semantics Aligning Network (SAN), which employed a decoder to reconstruct a dense semantics aligned full texture image. They supervise the ReID task and the semantic texture generating process simultaneously to learn a semantics-aligned feature representation. Quan \textit{et.al} \cite{RN448} employed the Neural Architecture Search (NAS) to find a part-aware network in a retrieval-based search space automatically. 

These methods above can be categorized as a one-branch network. These one-branch networks need to be robust to various challenging problems in the ReID task, which makes these networks overburdened. Although these methods utilized sophisticated designs or additional information to improve the performance, these endeavors make the one-branch network complicated and bring a limited improvement.

\subsection{Multi-Branch Networks}
There are many multi-branch networks for the ReID task \cite{RN427} \cite{RN426} \cite{RN428} \cite{RN335} \cite{RN339} \cite{RN430}. The first category uses a teacher network to teach a student network. Porrello \textit{et.al} \cite{RN415} introduced multiple Views Knowledge Distillation (VKD), which trains the teacher network using multiple views and only gives the student a small set of input views. After the knowledge distillation process, the student outperforms his teacher in the image-to-video setting. Zhuo \textit{et.al} \cite{RN427} employed a teacher-student framework for occluded ReID. They train the teacher network with a co-saliency network to simulate the occluded ReID, which enables the teacher to perceive the occlusion. Then they use the teacher network to generate the occluded mask to supervise the student network. These methods above employed a two-stage training process where only one network is trained in each stage. Therefore, they still under the paradigm of a one-branch network.

The second category trains each branch simultaneously and makes them co-teach. Yang \textit{et.al} \cite{RN428} proposed asymmetric co-teaching for the cross-domain ReID. They employed two branches and fed them with samples as pure as possible and as miscellaneous as possible, respectively. To achieve this goal, they encouraged their two networks to promote each other. Ge \textit{et.al} \cite{RN339} proposed Mutual Mean-Teaching (MMT) for the cross-domain ReID. They employed mean network, soft classification loss, and soft triplet loss to let two networks mutual-teach. The mean network is updated using the running average mean weight of each network.  Zhang \textit{et.al} \cite{RN187} proposed Deep Mutual Learning (DML) and gave two branches the same optimization objective. Although using one branch can complete this task, the DML scheme can find a much wider minimum for its loss function and provide a better generalization performance. However, each branch in these methods still needs to deal with various challenges, resulting in a limited performance improvement.

\subsection{Person Re-identification in A Specific Scene}
There are various challenging scenes in the ReID task. However, there are no benchmarks designed for the scale variation scene. On the other side, two large-scale benchmarks, P-DukeMTMC-reID and Occluded-DukeMTMC, are proposed for the occluded scenes recently. The occluded ReID has raised increasing attention from the computer vision community \cite{RN139} \cite{RN349} \cite{RN291} \cite{RN229} \cite{RN477} \cite{RN472}. In this field, Miao \textit{et.al} \cite{RN291} introduced Pose-Guided Feature Alignment (PGFA) and exploited pose landmarks to disentangle the useful information from the occlusion noise. However, this method largely depends on an accurate human pose estimator to detect human landmarks. Sun \textit{et.al} \cite{RN229} introduced Visibility-aware Part Model (VPM) and employed self-\\supervision learning to enable the model visibility-aware. Due to the limited self-supervision, the VPM learns a coarse division strategy, which limits its performance. 

These methods above merely focused on a specific scene and may fail to handle the ReID task in general scenes. On the contrary, our scheme is effective in various scenes. We divide the challenging scenes in the ReID task into multiple simpler ones and conquer them individually. Each branch achieves the perceptive ability for a particular scene in the training process. Concatenating features from each branch aggregate these different perceptive abilities and produce significant performance improvement.

\section{Learning to Disentangle Scenes} \label{method}
\begin{figure*}
	\centering
	\includegraphics[width=1\textwidth]{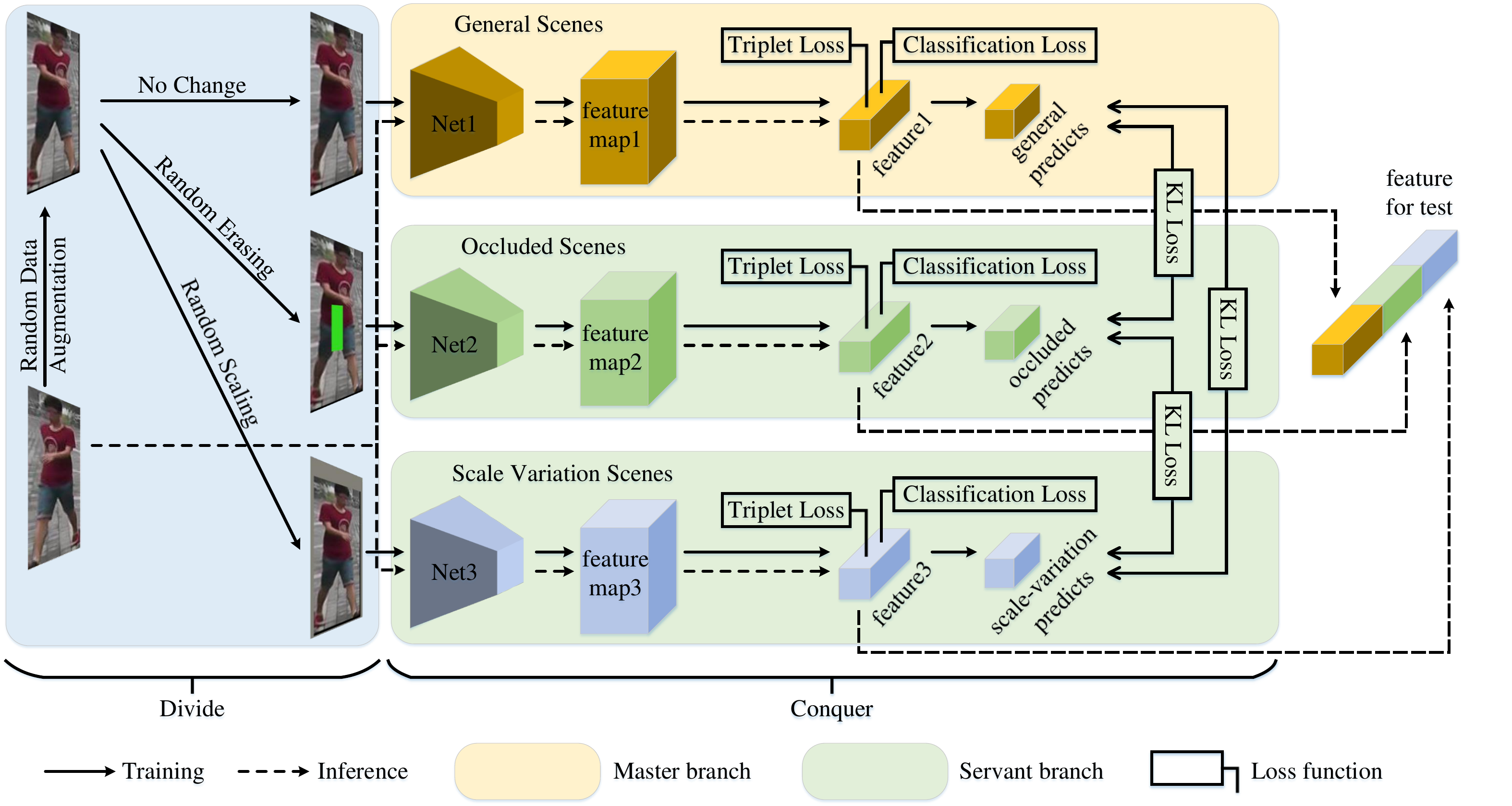}
	\caption{The framework of the proposed LDS method. We propose a divide-and-conquer strategy to deal with the ReID task. This framework contains two parts, ``divide'' and ``conquer'' parts. They are used to ``divide'' the complex scenes and ``conquer'' the specific scene. In the ``conquer'' part, we use mutual learning to promote each branch.}
	\label{framework}
\end{figure*}

This section elaborates on the proposed method for ReID, \textit{i.e.,} Learning to Disentangle Scenes (LDS). The framework of the proposed LDS method is illustrated in Fig. \ref{framework}. In this framework, we adopt the design philosophy of \textbf{divide-and-conquer} to deal with the ReID task.

\subsection{``Divide'' the Complex Scenes} \label{scenesimulation}
This paper identifies the occlusion and scale variation scenes from the complex scenes in the ReID task. We adopt a self-supervision operation to generate new images with the controlled characteristics. First, we apply a \textit{random data augmentation} strategy to each image. The \textit{random data augmentation} introduces more samples for the network training. Then we make three copies from the new samples.

\paragraph{Occlusion Scenes.} To generate an image with occlusion, we apply the \textit{random erasing} to the first copy. The probability of \textit{random erasing} is set to 1 to ensure occlusion exists in this image.

\paragraph{Scale Variation Scenes.} To generate an image with scale variation, we propose the \textit{random scaling} and apply it to the second copy. The \textit{random scaling} is described in detail below. We first generate a baseboard with the mean value of three channels (R, G, B) of all the images in ImageNet. Then we scale the second copy to 0.8 $\sim$ 1.1 times its original size. The zoom value is randomly generated. If the zoom value is less than 0.9, the scaled image is pasted in the baseboard center. For the ReID task, the center of input images is often informative, and the marginal part usually contains background and noise information. Putting it at the center of the baseboard makes the servant branch focus on the center of input images. If the zoom value is between 0.9 and 1.0, the scaled image is pasted anywhere on the baseboard. If the zoom value is more than 1.0, the scaled image is pasted at the baseboard center. All marginal parts beyond the baseboard boundary are discarded. The probability of the \textit{random scaling} operation is also set to 1. We set the minimum zoom value to 0.8 because a smaller margin around the image can improve the performance. Meanwhile, a much larger zoomed image introduces more information loss. Through a co-teach strategy in Section \ref{mutuallearning}, the master and servant branches focus on the image center, making each branch neglect the marginal part and improve the overall performance.

\begin{figure}[!h]
	\centering 
	\subfigure[Input]{	\includegraphics[scale=0.28]{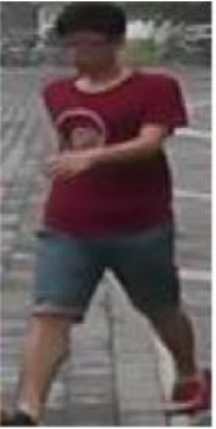}} 
	\subfigure[0.8 $\sim$ 0.9]{ \includegraphics[scale=0.28]{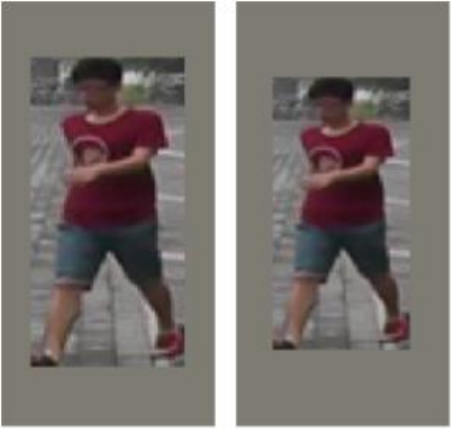}}  
	\subfigure[0.9 $\sim$ 1.0]{\includegraphics[scale=0.28]{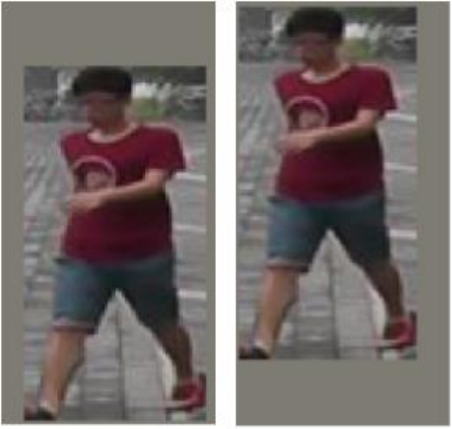}}  
	\subfigure[1.0 $\sim$ 1.1]{	\includegraphics[scale=0.28]{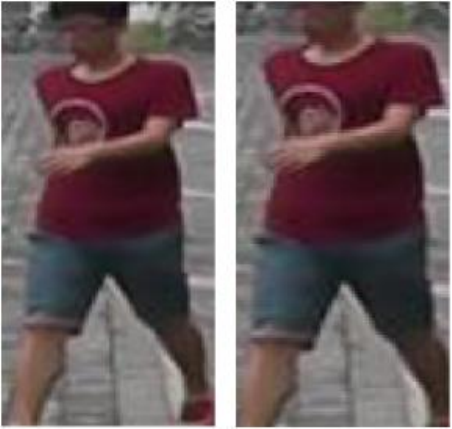}} 
	\caption{Random scaling. (a) Original input image. (b), (c), (d) Examples when the zoom scale in different ranges.} \label{randomscale}
\end{figure}

\paragraph{General Scenes.} We keep the third copy without any artificial change. The reason is explained below. The first and second copies are manipulated by the \textit{random erasing} and the \textit{random scaling}, respectively. Thus some useful information in them is lost. We keep the third copy as the original one to provide the missing information to the first and second ones. On the other side, the third copy represents the images that happened in general scenes.

\begin{figure}[!h]
	\centering
	\subfigure[Heterologous Input]{
		\begin{minipage}[b]{0.19\textwidth} \label{heterologous}
			\centering
			\includegraphics[width=1\textwidth]{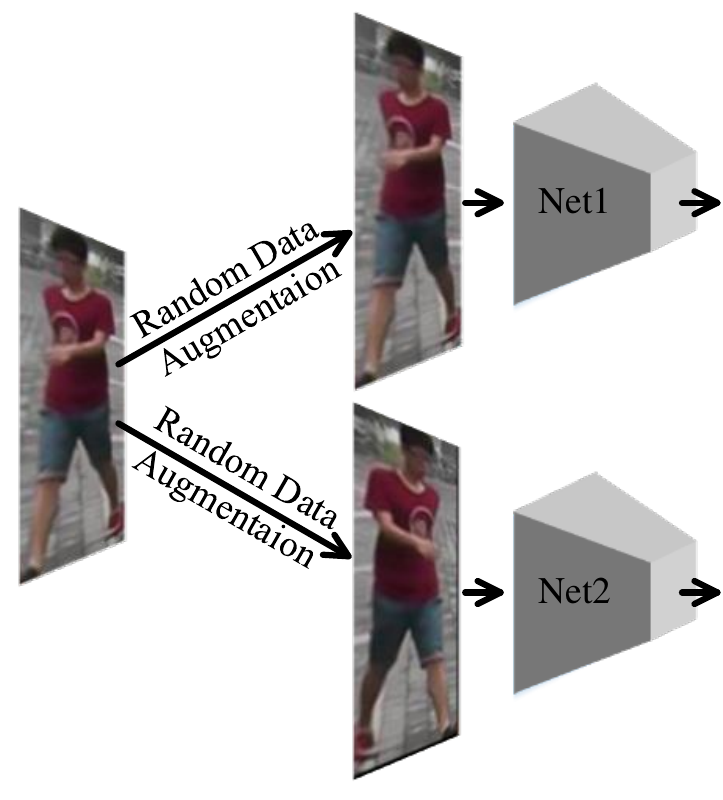}
	\end{minipage}          }
	\subfigure[Homologous Input]{
		\begin{minipage}[b]{0.25\textwidth} \label{homologous}
			\centering
			\includegraphics[width=1\textwidth]{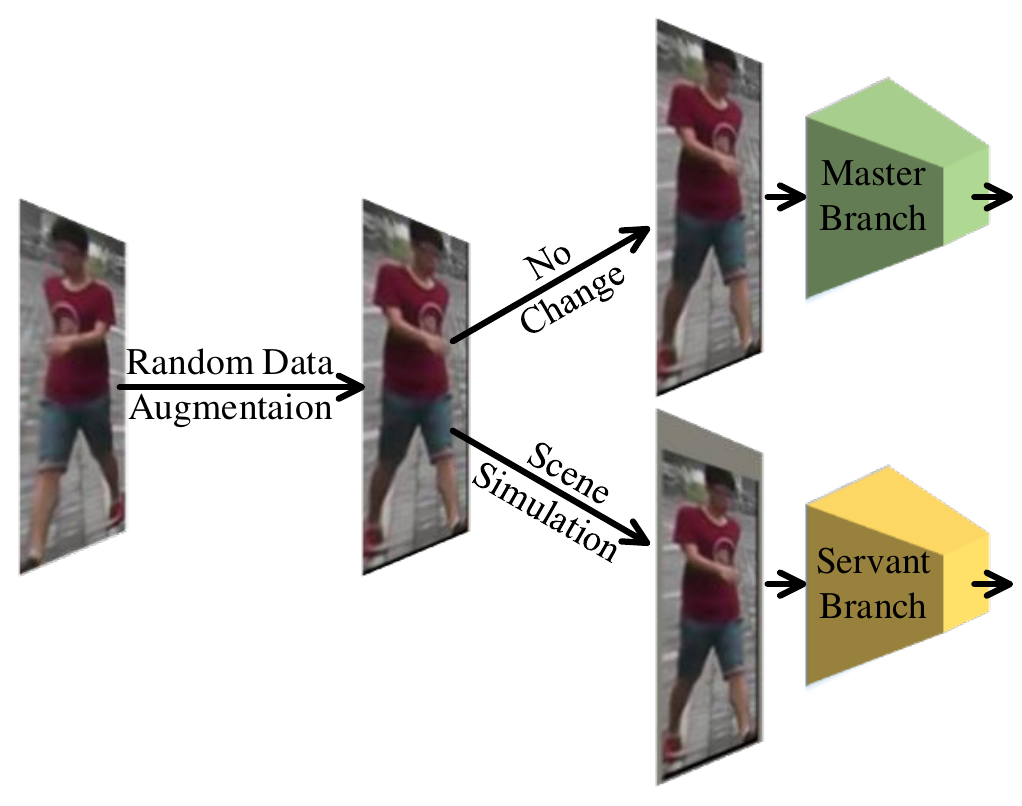}
	\end{minipage}          }
	\caption{Illustrations of \textit{heterologous input} and the proposed \textit{homologous input}. The \textit{homologous input} is applied to our scheme in Fig. \ref{framework}.} \label{inputs}
\end{figure}

\paragraph{Image Alignment for Different Scenes.} There are misalignment problems for the traditional multi-branch networks, as illustrate in Fig. \ref{heterologous}. This misalignment problem is mainly derived from the different input images, denoted as the \textit{heterologous input}. In general, the \textit{random data augmentation} operation usually includes \textit{random flipping} and \textit{random cropping}. These operations make the input images change their orientations and center positions, which makes the different branches receive misaligned images. Therefore, the \textit{heterologous input} utilized by most multi-branch networks often results in a misalignment problem. We propose \textit{homologous input} to solve this problem, as illustrated in Fig. \ref{homologous}. The self-supervision operation is after the \textit{random data augmentation}. This fashion ensures that the different branches have the same source image. This \textit{homologous input} is simple but effective, and the latter extensive ablation studies demonstrate its effectiveness.

\subsection{``Conquer'' the Specific Scene}
We propose a multi-branch network to ``conquer'' each specific scene. This network consists of three branches, including the master branch and two servant branches. The master branch deals with the ReID problem in general scenes, while servant branches handle the occluded scenes and scale variation scenes. We formulate this optimization process for each branch. Given $N$ image samples $\mathcal{I} = \{x_i\}_{i=1}^N$, there are corresponding person IDs as $\mathcal{Y} = \{y_i\}_{i=1}^N$ where $y_i \in \{1, 2, \cdots, M\}$. $M$ is the total number of person identities. After the process of \textit{homologous input}, the input image $I_i$ becomes three copes with different characteristics denoted as $I_i^{general}, I_i^{occlude}$, and $I_i^{scale}$, which are fed to the master branch, the servant branch for the occluded scenes, and the servant branch for the scale variation scenes, respectively. We employ $\Theta$ to represent each network and use $I_i^{\Theta}$ to represent the generated image, $I_i^{general}, I_i^{occlude}$, and $I_i^{scale}$, for each branch. The feature map extracted from one specific branch is denoted as $F_i^{\Theta}$. Then, each feature map is processed by average pooling and BN Neck layer to generate a normalized feature $f_i^{\Theta}$.

The first loss function for optimizing each branch is the triplet loss. We employ existing soft margin triplet loss with batch hard mining \cite{RN487}, which is calculated as follows:
\begin{equation}
	\begin{split}
		\mathcal{L}^{\Theta}_{triplet} = \frac{1}{B}
		\sum_{j=1}^B \sum_{a\in b_i}
		\ln{} 
		\{ 
		1 & + 
		\exp{}
		[  
		\xi   + \max_{p\in \mathcal{P}(a)} {d(f_a^{\Theta}, f_p^{\Theta})} \\
		& - \min_{n\in \mathcal{N}(a)} {d(f_a^{\Theta}, f_n^{\Theta})} 
		]     \},  
	\end{split} \label{triplet}
\end{equation}
\noindent where $b_i$ is the $i^{th}$ batch, $B$ is the number of batches in each benchmark, $a, p, n$ are anchor, positive, and negative samples, respectively, $\mathcal{P}(a)$ and $\mathcal{N}(a)$ are the positive and negative sample sets corresponding to the given anchor $a$ in this batch, $\xi$ is the distance margin threshold, and function $d(\bullet)$ calculates the Euclidean distance between two extracted features. In Eq. \ref{triplet}, we follow the previous research \cite{RN196} \cite{RN502} \cite{RN487} and replace the hinge function $[m + \bullet]_+$ with the soft-plus function $\ln{}(1+\exp{}(\bullet))$. The soft-plus function is a smooth approximation of the hinge function and decays exponentially without having a hard cut-off, resulting in a numerically stable implementation. 

The second loss function for optimizing each branch is the classification loss. We use the additive margin softmax (AM-softmax) \cite{RN486} to calculate this loss as follows:
\begin{equation}
	\begin{split}
		\mathcal{L}^{\Theta}_{cls} = -\frac{1}{N}\sum_{i=1}^N{\log{\frac{e^{[\gamma(W_{y_i}f_i^{\Theta}-m)]}}{
					e^{[\gamma(W_{y_i}f_i^{\Theta}-m)]}+\sum\limits_{j=1, j\neq y_i}^Me^{(\gamma W_j f_i^{\Theta})}}}}, 
	\end{split} \label{cls}
\end{equation} 

\noindent where $W_{y_i}$ and $W_j$ are the weight vectors associated with class $y_i$ and $j$ in the final classification layer, $\gamma$ is the scaling factor, and $m$ is the margin to distinguish the similarity distance. In Eq. \ref{cls}, we follow the \cite{RN504} \cite{RN505} \cite{RN506} \cite{RN486} to calculate the inner product using normalized weights vectors, $W_{y_i}$ and $W_j$, and normalized feature $f_i$. 
The scaling factor $\gamma$ is set to a constant instead of a learned weight to accelerate the training process. 

For each specific scene, the loss function $L^{\Theta}_{conquer}$ is formulated as follows: 
\begin{equation}
	\mathcal{L}^{\Theta}_{conquer} = \mathcal{L}^{\Theta}_{triplet} + \mathcal{L}^{\Theta}_{cls}.
\end{equation}

\subsection{Mutual Learning} \label{mutuallearning}
We employ mutual learning to make each branch communicate its knowledge to others. Since all the branches have the same source image, their logits should have a similar data distribution. The difference between their logits is mainly derived from the missing information introduced by the self-supervision operations. Through mutual learning, the servant branch becomes ``sensitive'' to the specific scene and can ``guess'' the missing information under the guidance of knowledge from the master branch. In this way, the servant branch gets the perceptive ability for the specific scene. Meanwhile, the master branch receives knowledge from more scenes, which reduces its overfitting probability and improves its robustness.

We use the Kullback Leibler (KL) Divergence to quantify the similarity of logits from different branches. We first calculate the probability $p^{\Theta}$ of class $m$ for pedestrian image sample $x_i$ as follows:
\begin{equation}
	p_m^{\Theta}(x_i) = \frac{\exp{(\gamma W_{m}f_i^{\Theta})}}{\sum\limits_{k=1}^M\exp{(\gamma W_k f_i^{\Theta})}},
\end{equation} \label{probability}

\noindent where $\gamma W_{m}f_i^{\Theta}$ is the logit fed to the ``softmax” layer in the branch $\Theta$. We optimize each branch by employing a KL loss which is calculated as follows:
\begin{equation}
	\mathcal{L}_{ML}^{\Theta} = \frac{1}{S-1} \sum_{s=1,s\neq \Theta}^S \sum_{i=1}^N\sum_{m=1}^M p_m^{\Theta}(x_i)\log{\frac{p_m^{\Theta}(x_i)}{p_m^{s}(x_i)}},
\end{equation} \label{KL}

\noindent where $S$ is the branch number. The KL loss $\mathcal{L}_{KL}^{\Theta}$ makes the logits from different branches as similar as possible. For each branch, the loss function $L^{\Theta}$ is formulated as follows: 
\begin{equation}
	\mathcal{L}^{\Theta} = \mathcal{L}^{\Theta}_{ML} + \mathcal{L}^{\Theta}_{conquer}.
\end{equation}

In this paper, we train the proposed LDS in an end-to-end manner. There are three losses for all branches, \textit{i.e.,} $L^{general}, L^{occlude}$ and $L^{scale}$. The overall optimization function for our scheme is calculated as follows,
\begin{equation}
	\mathcal{L} = \mathcal{L}^{general} + \mathcal{L}^{occlude} + \mathcal{L}^{scale}.
\end{equation}

\subsection{Advantages of Proposed LDS Method}
Existing works usually use a one-branch network for the challenging ReID task. There are many challenging problems that need to be dealt with, making the one-branch network overburdened. Many sophisticated designs and additional information are utilized to strengthen the one-branch network. However, these designs make the one-branch network complicated, and the performance improvement is limited.

The Knowledge Distillation (KD) \cite{RN507} is proposed to enable a small network to become strong through learning the knowledge from a large one. The difficult learning task is assigned to the teacher network. Then the teacher is fixed, and the knowledge is distilled to the student. Although having fewer parameters, the student becomes as strong as the teacher. However, only one network learns in each training stage. Thus these KD methods are still under the paradigm of a one-branch network.

Mutual learning or co-teaching methods, such as DML \cite{RN187}, MMT \cite{RN339}, make two same networks share the responsibility and promote each other. The performance improvement is mainly due to a different network weight initialization. This operation makes each branch learn in different directions, and the KD loss makes them have an intermediate and better optimization direction. The multi-branch network often achieves better performance than a one-branch network. However, the number of challenging problems is not reduced, and each branch in these methods has the same responsibility, leading to difficulty in improving performance.

The proposed LDS introduces a divide-and-conquer strategy for the ReID task. Through self-supervision operations, the challenges are divided into simpler ones. Each servant branch only needs to deal with a specific challenge, which reduces the burden of each branch. By employing mutual learning, the servant branches can receive the knowledge from the master branch and obtain the ability to recover the missing information. The knowledge from the servant branches contains incomplete information, which reduces the overfitting possibility for the master branch. Therefore, the performance improvement of the proposed LDS is from the knowledge communication of different scenes. Correspondingly, the performance improvement of the traditional co-teaching methods is from the different initial conditions, \textit{i.e.,} random initialization of network weights. 

On the other side, the proposed LDS is a general and flexible framework. More image transformation operations can be introduced to deal with other challenging issues, \textit{i.e.,} lighting variation, similar appearance, etc. This paper focuses on the divide-and-conquer strategy, and two scenes can illustrate the effects of this strategy. Therefore, we only investigate two typical scenes, occlusion and scale variation. Experiment results in the latter section demonstrate that the proposed LDS is more effective than the existing one-branch and multi-branch networks.

\section{Experiments} \label{Experiments}

\subsection{Benchmarks and Evaluation Metrics}
We evaluate the proposed LDS on three image-based ReID bechmarks, including Market1501 \cite{RN149}, DukeMTMC-reID \cite{RN189}, MSMT17 \cite{RN172}, and two occluded ReID benchmarks, P-DukeMTMC-reID \cite{RN475} and Occluded-DukeMTMC \cite{RN291}. 

The Market1501 contains 1501 person identities captured by six different cameras on campus. In the training set, 12936 images for 751 persons are used. There are 19732 and 3368 images of the rest 750 person identities for gallery and query in the testing set.

The DukeMTMC-reID benchmark consists of 1812 person identities collected by eight synchronized cameras from campus. There are 16522 images of 702 identities in the training set. There are 17661 and 2228 images of the other 702 identities for gallery and query in the test set.

The MSMT17 contains 4101 person identities captured by a 15-camera network, including 12 outdoor and 3 indoor cameras on campus. The training set includes 32621 images of 1041 identities. There are 82161 and 11659 images of the other 3060 identities for the gallery and query in the test set.

In P-DukeMTMC-reID, there are 12927 images of 665 person identities for training, including 2647 images with occlusion and 10280 images without occlusion. There are 11216 images from 634 person identities for test, including 2163 images with occlusion for query and 9053 images without occlusion forming the gallery.

The Occluded-DukeMTMC contains 15618 images of 702 person identities in the training set, 17661 images of 1110 person identities in the gallery, and 2210 images of 519 person identities in the query.

The Cumulative Matching Characteristics (CMC) \cite{RN168} and mean Average Precision (mAP) \cite{RN149} are reported. We use the Rank-$k$ scores to represent the CMC curve. All the experiments are performed in a single query setting.

\subsection{Implementation Details}
We use the PyTorch toolbox, FastReID \cite{RN489}, to achieve the proposed LDS. Additionally, we use the ResNet-ibn \cite{RN73} \cite{RN490} as our backbone and initialize it by the ImageNet \cite{RN396} pre-trained model. The non-local layer \cite{RN161} is also employed in our backbone. Each person image is resized to 384$\times$128. We set the batch size to 64 and use Adam \cite{RN511} with initialized learning rate $3.5\times10^{-4}$ to train each benchmark for 60 epochs. We use the cosine annealing part of the SGDR \cite{RN512} to adjust the learning rate. We also freeze the backbone in the first 2000 iterations for each benchmark to train the network. Then we train the whole multi-branch network for the rest iterations. 

\begin{table*}
	\centering
	\begin{threeparttable}
		\centering
		\caption{Performance comparisons with state-of-the-art methods on Market1501, DukeMTMC-reID, and MSMT17. } 
		\label{SOTA}
		\begin{tabular}{llccccccc}
			\hline
			\multicolumn{1}{l}{\multirow{2}{*}{}} & \multicolumn{1}{l}{\multirow{2.5}{*}{Method}} & \multicolumn{1}{c}{\multirow{2.5}{*}{Publication}} & \multicolumn{2}{c}{Market1501} & \multicolumn{2}{c}{DukeMTMC-reID} & \multicolumn{2}{c}{MSMT17} \\
			\cmidrule(r){4-5} \cmidrule(r){6-7} \cmidrule(r){8-9}
			\multicolumn{1}{c}{} & \multicolumn{1}{c}{} & \multicolumn{1}{c}{}    & R1       & mAP   & R1       & mAP & R1       & mAP   \\ \hline
			\multirow{9}{*}{\begin{tabular}[c]{@{}c@{}}Attention-\\based\end{tabular}} 
			& BAT \cite{RN480}                 & ICCV19       & 95.10    & 87.40   & 87.70   & 77.30  &  79.50  & 56.80 \\  
			& ABD-Net \cite{RN407}             & ICCV19       & 95.60    & 88.28   & 89.00   & 78.59  &  82.30  & 60.80 \\  
			& CAR \cite{RN515}                 & ICCV19       & 96.10    & 84.70   & 86.30   & 73.10  &         &       \\  
			& SCAL (spatial) \cite{RN517}      & ICCV19       & 95.40    & 88.90   & 89.00   & 79.60  &         &       \\  
			& SONA$^{2+3}$-Net$\mu$ \cite{RN520} & ICCV19     & 95.58    & 88.83   & 89.38   & 78.23  &         &       \\  
			& MHN-6 (PCB) \cite{RN289}         & ICCV19       & 95.10    & 85.00   & 89.10   & 77.20  &         &       \\  
			& IANet \cite{RN481}               & CVPR19       & 94.40    & 83.10   & 87.10   & 73.40  &  75.50  & 46.80 \\  
			& SCSN (3 stage) \cite{RN484}      & CVPR20       & 95.70    & 88.50   & 90.10   & 79.00  &  83.00  & 58.00 \\  
			& RGA-SC \cite{RN405}              & CVPR20       & 96.10    & 88.40   &         &        &  80.30  & 57.50 \\  \hline 
			\multirow{6}{*}{\begin{tabular}[c]{@{}c@{}}Semantics-\\based\end{tabular}}
			& $P^2$-Net (+triplet loss) \cite{RN456} & ICCV19 & 95.20    & 85.60   & 86.50   & 73.10  &         &       \\  
			& DSA-reID \cite{RN519}            & CVPR19       & 95.70    & 87.60   & 86.20   & 74.30  &         &       \\  
			& SAN \cite{RN345}                 & AAAI20       & 96.10    & 88.00   & 87.90   & 75.50  &  79.20  & 55.70 \\  
			& DLBC \cite{RN516}                & ACM MM20     & 94.60    & 87.40   & 88.70   & 78.50  &  78.20  & 55.60 \\  
			& ISP \cite{RN452}                 & ECCV20       & 95.30    & 88.60   & 89.60   & 80.00  &         &       \\  \hline 
			\multirow{4}{*}{\begin{tabular}[c]{@{}c@{}}Stripe/Part-\\related\end{tabular}}
			& Auto-ReID \cite{RN448}           & ICCV19       & 94.50    & 85.10   &         &        &  78.20  & 52.50 \\  
			& BDB + Cut \cite{RN468}           & ICCV19       & 95.30    & 86.70   & 89.00   & 76.00  &         &       \\  
			& RRID \cite{RN314}                & AAAI20       & 95.20    & 88.90   & 89.70   & 78.60  &         &       \\  
			& HAA \cite{RN460}                 & ACM MM20     & 95.80    & 89.50   & 89.00   & 80.40  &         &       \\  \hline 
			\multirow{8}{*}{Others}
			& SFT \cite{RN457}                 & ICCV19       & 93.40    & 82.70   & 86.90   & 73.20  &  73.60  & 47.60 \\  
			& DCDS \cite{RN450}                & ICCV19       & 94.81    & 85.80   & 87.50   & 75.50  &         &       \\  
			& VCFL \cite{RN518}                & ICCV19       & 89.25    & 74.48   &         &        &         &       \\  
			& MVP Loss \cite{RN521}            & ICCV19       & 91.40    & 80.50   & 83.40   & 70.00  &  71.30  & 46.30 \\  
			& OSNet \cite{RN479}               & ICCV19       & 94.80    & 84.90   & 88.60   & 73.50  &  78.70  & 52.90 \\  
			& DSFL \cite{RN482}                & ACM MM20     & 96.20    & 89.90   & 90.20   & 81.10  &  84.20  & 60.70 \\  
			& NEWTH \cite{RN483}               & NeurIPS20    & 95.60    & 89.40   &         &        &  71.50  & 53.10 \\  
			& M$^3$ + HA-CNN \cite{RN522}      & CVPR20       & \textbf{96.50}  & 85.20 & 87.10 & 72.20 & 74.30 & 43.80 \\  
			& CtF \cite{RN454}                 & ECCV20	      & 93.70    & 84.90   & 87.60   & 74.80  &         &       \\  \hline 
			\multirow{5}{*}{multi-branch}
			& DML \cite{RN187}				   & CVPR18       & 89.34    & 70.51   &         &        &         &      \\
			& PTL + MGN \cite{RN491}           & IJCAI19      & 94.83    & 87.34   & 89.36   & 79.16  &  73.12  & 41.38 \\  
			& CAMA (N=3) \cite{RN514}          & CVPR19       & 94.70    & 84.50   & 85.80   & 72.90  &         &       \\  
			& HBFP-Net \cite{RN513}			 & ACM MM20       & 95.80    & 89.80   & 89.50   & 80.20  &         &       \\  
			& Proposed LDS           &                        & 95.84    & \textbf{90.37} & \textbf{91.56} & \textbf{82.50} & \textbf{86.54} & \textbf{67.21} \\  \hline \hline
			\multirow{6}{*}{+ Re-rank}
			& VCFL \cite{RN518} + Re-rank        & ICCV19       & 90.91    & 86.67  &        &        &         &       \\
			& DCDS \cite{RN450} + Re-rank        & ICCV19       & 95.40    & 93.30  & 88.50  & 86.10  &         &       \\
			& Auto-ReID \cite{RN448} + Re-rank   & ICCV19       & 95.40    & 94.20  &        &        &         &       \\
			& SFT \cite{RN457} + Re-rank         & ICCV19       & 93.50    & 90.60  & 88.30  & 83.30  &  76.10  & 60.80 \\
			& MVP Loss \cite{RN521} + Re-rank    & ICCV19       & 93.30    & 90.90  & 86.30  & 83.90  &         &       \\ 
			& Proposed LDS + Re-rank             &  & \textbf{96.17} & \textbf{94.89} & \textbf{92.91} & \textbf{91.00} & \textbf{88.35} & \textbf{79.09}\\ \hline
		\end{tabular}
		\begin{tablenotes}
			\item[1] The best results are in bold. 
			\item[2] The metric `R1' is the abbreviation of `Rank-1'.
		\end{tablenotes}
	\end{threeparttable}
\end{table*}

\subsection{Comparison with State-of-the-Arts}
Table \ref{SOTA} represents the performance comparisons between the proposed LDS and other state-of-the-art methods on three popular benchmarks in terms of CMC accuracy and mAP scores. These methods are within two yeas and include nine attention-based methods, six semantics-based methods, four stripe/part-related methods, three multi-branch networks, and nine other kinds of methods. They are all trained on the standard training sets without depending on additional images or labels. Early literature often used the re-rank \cite{RN523} technique. This technique can effectively adjust the order of image candidates and improve the mAP scores. In Table \ref{SOTA}, we also present the performance of the proposed LDS with the re-rank.

\paragraph{Performances on Market1501.}
In table \ref{SOTA}, compared to the other state-of-the-art methods, the proposed LDS achieves competitive results. There are other three multi-branch methods, PTL \cite{RN491}, CAMA \cite{RN514}, and HBFP-Net \cite{RN513}. These methods employed the feature map from different layers to form a rich feature representation. Meanwhile, our method gives each branch a different perceptive ability by feeding them images with different characteristics. Thus, the proposed LDS is more easily implemented. And LDS also achieves a better performance than them. The proposed LDS with re-rank also achieves better performance. These extensive comparisons demonstrate the effectiveness of our scheme.

\paragraph{Performances on DukeMTMC-reID.}
Table \ref{SOTA} shows that the proposed LDS achieves state-of-the-art performance. Compared to the best competitor, DSFL \cite{RN482}, our method achieves performance improvement of 1.3\% and 1.4\% on the metric of Rank-1 and mAP, respectively. For the re-rank counterparts, we conduct performance improvement of 4.4\% and 4.7\% on the metric of Rank-1 and mAP compared to the best competitor, DCDS \cite{RN450}. These comparisons demonstrate our scheme achieves considerable improvement compared to other state-of-the-art methods.

\paragraph{Performances on MSMT17.}
Table \ref{SOTA} also represents the proposed LDS achieves the state-of-the-art performance. For the metric of Rank-1, we achieve a performance improvement of 2.3\% compared to the best competitor, DSFL \cite{RN482}. For the metric of mAP, we achieve a performance improvement of 6.4\% compared to the best competitor, ABD-Net \cite{RN407}. For the re-rank version, our scheme achieves significant improvement in the metric of mAP, \textit{i.e.,} 18.29\%, compared to the best competitor, SFT \cite{RN457}. These comparisons demonstrate the effectiveness of our method.

\subsection{Ablation Study}
Table \ref{ablation} shows the ablation study. The baseline is a one-branch network and trained using classification loss and triplet loss, following the same backbone and training parameters with the proposed LDS. We also use DML \cite{RN187} as the baseline of the multi-branch network. The evaluation of DML also uses the concatenated features from different branches. In table \ref{ablation}, the LDS-$i^{(j)}$ denotes the $j^{th}$ configuration of the LDS with $i$ branches, and the DML-$i$ denotes the DML with $i$ branches.

In Table \ref{ablation}, the proposed \textit{homologous input} ensures the different branches have the same source image. The LDS with two branches has one master branch and one servant branch, and the LDS with three branches has one master branch and two servant branches. These servant branches are utilized to deal with occluded and scale variation scenes.

\begin{table*}[t]
	\centering
	\begin{threeparttable}
		\caption{Performance comparisons with the baseline and DML. }
		\label{ablation}
		\begin{tabular}{llccccccccc}
			\hline
			\multirow{2.5}{*}{\# Branch} & \multirow{2.5}{*}{Method} & \multirow{2.5}{*} {\begin{tabular}[c]{@{}c@{}}Random\\ Erasing\end{tabular}} & \multirow{2.5}{*} {\begin{tabular}[c]{@{}c@{}}Random\\ Scaling\end{tabular}} & \multirow{2.5}{*} {\begin{tabular}[c]{@{}c@{}}Homologous\\ Input\end{tabular}} & \multicolumn{2}{c}{Market1501} & \multicolumn{2}{c}{DukeMTMC-reID} & \multicolumn{2}{c}{MSMT17} \\
			\cmidrule(r){6-7} \cmidrule(r){8-9} \cmidrule(r){10-11}
			&         &        &    &    & R1       & mAP     & R1       & mAP    & R1       & mAP    \\ \hline
			1-Branch                  & Baseline    &     &   &         & 95.01     & 86.57    & 88.78    & 76.29    & 81.65     & 55.53   \\ \hline
			\multirow{6}{*}{2-Branch} 
			& DML-2\cite{RN187}\dag &              &              &                           & 95.34  & 87.76    & 89.59     & 77.91   & 84.30  & 60.30   \\
			& LDS-2$^{(1)}$    &              &              & $\checkmark$              & 95.55  & 88.28    & 90.44     & 79.21   & 84.92  & 61.62   \\
			& LDS-2$^{(2)}$    &              & $\checkmark$ &                           & 95.37  & 87.75    & 89.00     & 77.80   & 83.94  & 59.58   \\
			& LDS-2$^{(3)}$    &              & $\checkmark$ & $\checkmark$              & 95.61  & 88.19    & 89.99     & 79.31   & 84.73  & 61.35   \\ 
			& LDS-2$^{(4)}$    & $\checkmark$ &              &                           & 95.75  & 89.94    & 90.93     & 81.74   & 86.10  & 66.24   \\
			& LDS-2$^{(5)}$    & $\checkmark$ &              & $\checkmark$              & 95.55  & 90.24    & 90.98     & 82.17   & 86.49  & 67.05   \\ \hline
			\multirow{4}{*}{3-Branch} 
			& DML-3\cite{RN187}\dag &              &              &                           & 95.58  & 87.95    & 89.18     & 78.01   & 84.39  & 60.55   \\
			& LDS-3$^{(1)}$    &              &              & $\checkmark$              & 95.16  & 88.25    & 89.95     & 79.55   & 85.16  & 62.46   \\
			& LDS-3$^{(2)}$    & $\checkmark$ & $\checkmark$ &                           & 95.72  & 89.76    & 90.31     & 80.88   & 85.89  & 65.25   \\
			& LDS-3$^{(3)}$    & $\checkmark$ & $\checkmark$ & $\checkmark$              & \textbf{95.84}    & \textbf{90.37 }   & \textbf{91.56}    & \textbf{82.50}  & \textbf{86.54}   & \textbf{67.21}   \\ \hline
		\end{tabular}
		\begin{tablenotes}
			\item[1] '\dag': reimplemented by us.
			\item[2] The metric `R1' is the abbreviation of `Rank-1'.
		\end{tablenotes}
	\end{threeparttable}
\end{table*}

\paragraph{Effectiveness of Proposed Scheme.}
We make three comparisons to demonstrate the effectiveness of the proposed scheme. These comparisons are between the baseline and the LDS-2$^{(5)}$, between the DML-2 and the LDS-2$^{(5)}$, between the DML-3 and the LDS-3$^{(3)}$. Compared to the baseline, the LDS-2$^{(5)}$ configuration achieves noticeable performance improvement, \textit{i.e.,} an average Rank-1 improvement of 2.5\% and an average mAP improvement of 7\% for the three benchmarks. Compared to the DML-2 \cite{RN187}, the LDS-2 $^{(5)}$ has an average Rank-1 score of 1.2\% and an average mAP score of 4.4\% superiority. Compared to DML-3, the LDS-3$^{(3)}$ achieves an average Rank-1 score of 1.5\% and an average mAP score of 4.5\% performance improvement. These comparisons demonstrate the effectiveness of the proposed scheme over the one-branch and multi-branch networks. 

\paragraph{Effectiveness of Proposed Random Scaling.}
We make another three comparisons to demonstrate the effectiveness of the proposed \textit{random scaling}. These comparisons are between the baseline and the LDS-2$^{(3)}$, between the DML-2 and the LDS-2$^{(3)}$, between the LDS-2$^{(5)}$ and the LDS-3$^{(3)}$. Compared to the baseline, the LDS-2$^{(3)}$ configuration achieves an average Rank-1 improvement of 1.6\% and average mAP improvement of 3.4\% for the three benchmarks. Compared to the DML-2, the LDS-2$^{(3)}$ improves an average Rank-1 score of 0.3\% and an average mAP score of 0.9\%. Compared to the LDS-2$^{(5)}$, LDS-3$^{(3)}$ achieves an average Rank-1 score of 0.3\% and an average mAP score of 0.2\% performance improvement. These comparisons demonstrate the proposed \textit{random scaling} can effectively improve the performance. 
We also make another two comparisons, which are between the LDS-2$^{(2)}$ and the DML-2, between the LDS-2$^{(3)}$ and the DML-2. The LDS- 2$^{(2)}$ achieves inferior performance compared to the DML-2. Although introducing more scale variations, \textit{random scaling} introduces an additional misalignment problem when zoom value is between 0.9 and 1.0. In this situation, the misalignment problem gets severer, which results in a worse performance. After applying the \textit{homologous input}, the LDS-2$^{(3)}$ achieves better performance than the DML-2.

\paragraph{Effectiveness of Proposed Homologous Input.}
We make two comparisons to demonstrate the effectiveness of the proposed \textit{homologous input}. These comparisons are between the DML-2 and the LDS-2$^{(1)}$, between the DML-3 and the LDS-3$^{(1)}$. Compared to the DML-2, the LDS-2$^{(1)}$ employs the \textit{homologous input} and achieves an average Rank-1 score of 0.5\% and an average mAP score of 1.0\% performance improvement on the three benchmarks. The comparisons between the DML-3 and LDS-3$^{(1)}$ also have similar performance. We explain these performance improvements below. In the DML, the different optimization processes of each branch are mainly derived from the random initialization of the non-local layer and the \textit{random data augmentation}. The proposed \textit{homologous input} makes each branch have the same source images and avoids the misalignment problem introduced by the \textit{random data augmentation}. The random initialization of the non-local layer ensures the branches have a different learning process. Therefore, employing the \textit{homologous input} can help to achieve a noticeable performance improvement. 

\paragraph{Random Erasing vs. Proposed Random Scaling.}
Applying different servant branches brings different performance improvements. For the two-branch version in the table \ref{ablation}, applying servant branch for occlusion, \textit{i.e., } LDS-2$^{(5)}$ achieves better performance than scale variation, \textit{i.e.,} LDS-2$^{(3)}$. This phenomenon may be caused by the fact that the occlusion scenes are more common than the scale variation scenes in the three benchmarks. Or the \textit{random erasing} can increase the diversity of input images more effectively than the proposed \textit{random scaling}.

\paragraph{Effectiveness in Occluded Scenes.}
To verify effectiveness of the proposed scheme in the occluded scenes, we train LDS-2$^{(5)}$ on the P-DukeMTMC-reID and the Occluded-\\DukeMTMC benchmarks. These two large-scale benchmarks include training sets for model learning. The results are illustrated in table \ref{ablationP} and table \ref{ablationO}. The baseline method and the DML with two branches are also trained on these two benchmarks. The performances on these two large-scale benchmarks demonstrate the effectiveness of the proposed scheme in the occluded scenes. In table \ref{ablationP}, the proposed LDS 2$^{(5)}$ achieves state-of-the-art performance on P-DukeMTMC-reID benchmark under a supervised setting. In table \ref{ablationP}, PCB \cite{RN191} is a stripe/part-related method, and PVPM \cite{RN472} utilized the pose-guided attention to mine the part visibility for the occluded ReID task. Based on the strong baseline method, the proposed LDS 2$^{(5)}$ improves the performance further. In table \ref{ablationO}, the proposed LDS 2$^{(5)}$ achieves state-of-the-art performance on Occluded-DukeMTMC benchmark under a supervised setting. In table \ref{ablationO}, PGFA \cite{RN291} exploited the pose landmarks to disentangle the visible region from the occlusion noise. HOReID \cite{RN467} utilized the key-point information to obtain a local-feature graph to learn the high-order relation and topology knowledge. Compared with them, our method employs a simple idea and also \\achieves a better performance.

\begin{table}[!htpb]
	\centering
	\caption{Performance comparisons on the P-DukeMTMC-reID benchmark under a supervised setting. }
	\label{ablationP}
	\resizebox{\columnwidth}{!}{
		\begin{tabular}{llcccc}
			\hline
			Method       & Venue & Rank-1 & Rank-5 & Rank-10 & mAP  \\ \hline
			PCB \cite{RN191} & ECCV2018& 79.4  & 87.1 & 90.0 & 63.9 \\
			PVPM \cite{RN472} & CVPR2020 & 85.1   & 91.3   & 93.3    & 69.9 \\
			Baseline     & & 88.2   & 93.1   & 94.3    & 76.4 \\
			DML-2          & & 90.5   & 94.1   & 95.1    & 78.8 \\
			LDS-2$^{(5)}$ & & \textbf{91.9}   & \textbf{95.2}   & \textbf{96.3}    & \textbf{82.9} \\ \hline
	\end{tabular} }
\end{table}

\begin{table}[h]
	\centering
	\caption{Performance comparisons on the Occluded-DukeMTMC benchmark under a supervised setting. }
	\label{ablationO}
	\resizebox{\columnwidth}{!}{
		\begin{tabular}{llcccc}
			\hline
			Method       & Venue & Rank-1 & Rank-5 & Rank-10 & mAP  \\ \hline
			PGFA \cite{RN291} & ICCV2019& 51.4  & 68.6 & 74.9 & 37.3 \\
			HOReID \cite{RN467} & CVPR2020 & 55.1   &    &      & 43.8 \\
			Baseline     & & 62.6   & 75.1   & 80.6    & 50.2 \\
			DML-2          & & 63.6   & 76.4   & 80.5    & 51.9 \\
			LDS-2$^{(5)}$ & & \textbf{64.3}   & \textbf{77.1}   & \textbf{82.6}    & \textbf{55.7} \\ \hline
	\end{tabular} }
\end{table} 

\begin{table}[H]
	\centering
	\caption{Performance comparisons between the proposed LDS using mutual learning and master-servant learning on different benchmarks. }
	\label{ablationM}
	\resizebox{\columnwidth}{!}{
		\begin{tabular}{llccccccccc}
			\hline
			& \multicolumn{2}{c}{Market1501} & \multicolumn{2}{c}{DukeMTMC-reID} & \multicolumn{2}{c}{MSMT17} \\
			\cmidrule(r){2-3} \cmidrule(r){4-5} \cmidrule(r){6-7}
			& R1     & mAP      & R1        & mAP     & R1     & mAP    \\ \hline
			DML-3 \cite{RN187}         & 95.58  & 87.95    & 89.18     & 78.01   & 84.39  & 60.55   \\
			w/ MS Learning   & 95.64  & 89.89    & 91.34     & 81.97   & 86.27  & 66.25   \\ 
			w/ Mutual Learning           & 95.84  & 90.37    & 91.56     & 82.50   & 86.54  & 67.21   \\ \hline
	\end{tabular} }
\end{table} 

\paragraph{Effectiveness of Mutual Learning.}
Although multiple \\branches in our scheme employ mutual learning to transfer knowledge to each branch, we also propose master-servant learning (MS Learning). In MS learning, the knowledge communication is only between the master branch and servant branch. And there is no knowledge communication between the different servant branches. In table \ref{ablationM}, the performance comparisons between the proposed LDS using mutual learning and MS learning are listed. The LDS with MS learning achieves better performance than the DML-3. However, the LDS with MS learning is inferior to the LDS with mutual learning. We explain these performances below. For master-servant learning, the knowledge from different servant branches is indirectly communicated through the intermediate master branch. For mutual learning, the servant branches have direct communication, and the experiment results indicate this direct communication is more effective. Therefore, we apply mutual learning in our scheme to promote each branch.

\begin{figure}[!h]
	\centering
	\begin{tabular}[c]{cc}
		\multirow{10.6}{*}{ \subfigure[Query]{ \includegraphics[scale=0.33]{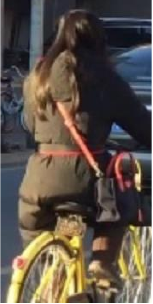} }	}
		& 		\subfigure[Baseline]{ \includegraphics[scale=0.33]{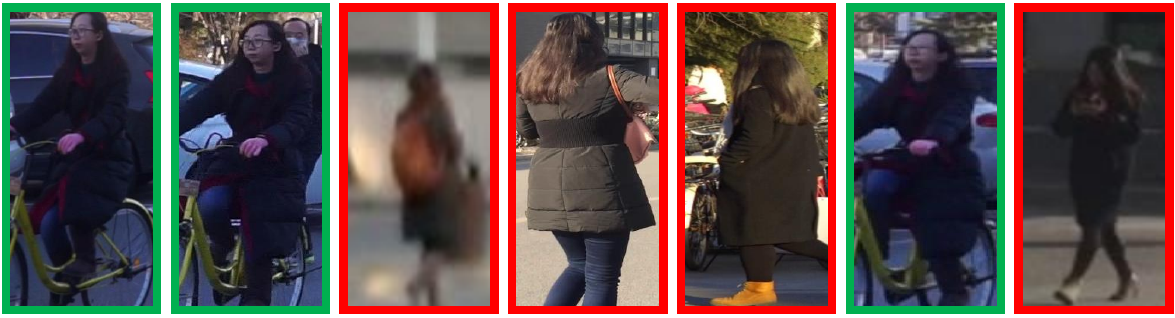} } \\ 
		&		\subfigure[DML]{	\includegraphics[scale=0.33]{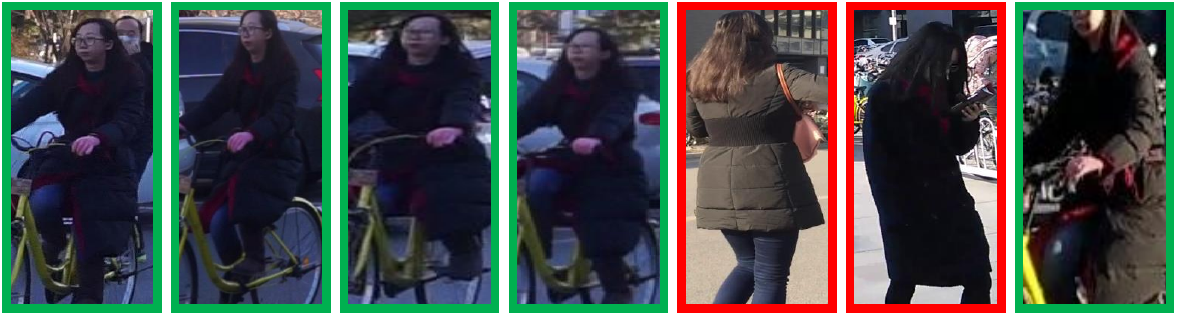}		} \\ 
		&		\subfigure[Proposed LDS]{	\includegraphics[scale=0.33]{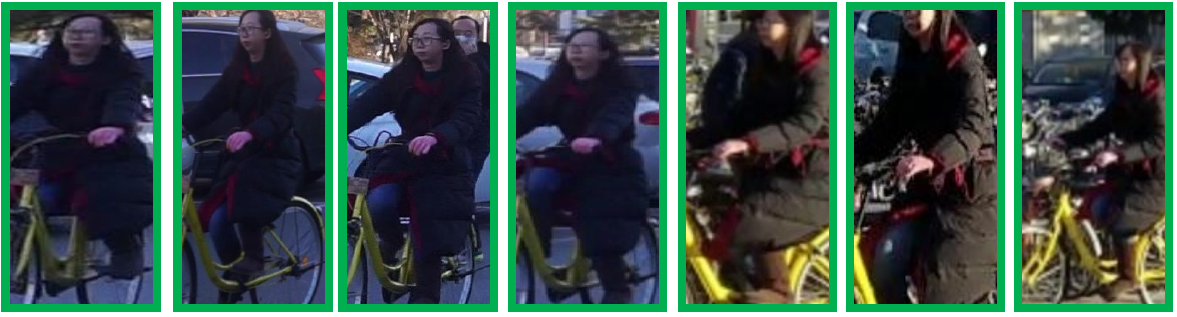}\label{fig:lds}		} \\ 
	\end{tabular}    
	\caption{Retrieval image examples. The query image is from the MSMT17 benchmark. The query and retrievals contain obvious occlusion and scale variation. The correct and incorrect ones are in a green and red box, respectively.}
	\label{fig:retrievalResults}
\end{figure}

\begin{figure}[!h]
	\centering
	\subfigure[Input]{ \includegraphics[width=0.08\textwidth]{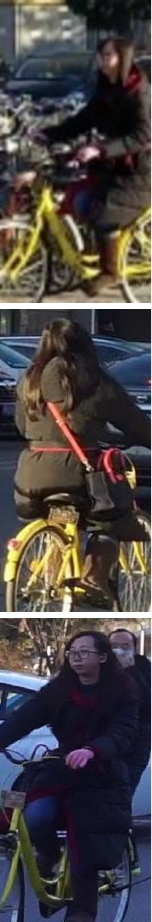} }
	\subfigure[Baseline]{ \includegraphics[width=0.08\textwidth]{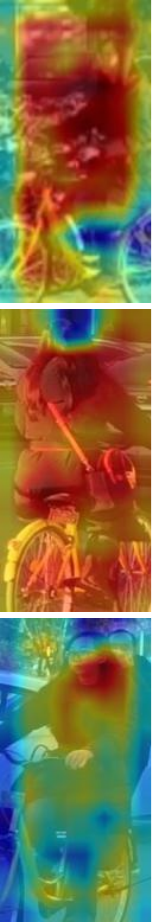} }
	\subfigure[DML]{ \includegraphics[width=0.08\textwidth]{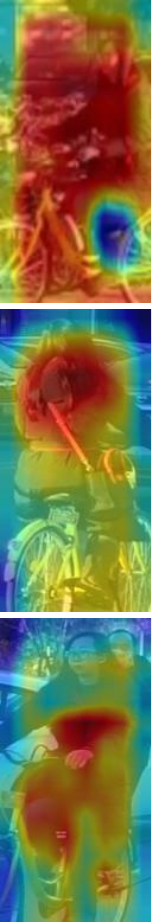} }
	\subfigure[Proposed LDS]{ \includegraphics[width=0.08\textwidth]{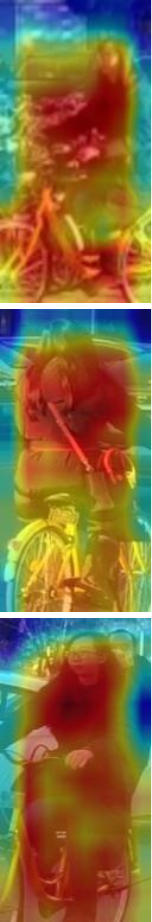} }
	\caption{Activation maps of images in Fig. \ref{fig:retrievalResults}. In each row, the first one is the original input image. The second, third, and fourth ones are the grad-CAM++ \cite{RN524} results from the models of baseline method, DML, and the proposed LDS, respectively.} \label{fig:cam}
\end{figure}

\paragraph{Qualitative Analysis.}
We show the visual comparisons of retrieval images in Fig. \ref{fig:retrievalResults}. We select a query image with occlusion from the MSMT17 benchmark and compare the retrievals of different methods, including baseline, DML, and the proposed LDS. In Fig. \ref{fig:retrievalResults}, the baseline method neglects many obvious correct results. The DML method identifies those easy image samples. However, it fails in the scale variation scenes, \textit{e.g.,} the last result in \ref{fig:lds}. These results demonstrate the effectiveness of our scheme in occluded scenes and scale variation scenes. To further analyze the learning ability of different models, we show the activation maps of different methods for the same input image, as illustrated in Fig. \ref{fig:cam}. Since the DML and the proposed LDS have three branches, we only show the activation map of the first branch. For each example, the activation map of the proposed LDS shows better responses than the others. These comparisons explain why the proposed LDS performs better and demonstrate its effectiveness.

\section{Conclusion} \label{conlusion}
In this paper, we propose to learn to disentangle scenes for the ReID task. This scheme employs a divide-and-conquer strategy for the ReID task. Concretely, we use two self-supervision operations to generate new image with the characteristics of occluded and scale variation scenes. Then we utilize two servant branches to deal with them. In this way, the burden of each branch is relieved. We also use a master branch to handle the general scenes. Mutual learning is employed to promote each branch. Through collaborative learning, the servant branch learns the missing information through guidance from the master branch. Moreover, the knowledge from the servant branches makes the master branch more robust. Extensive experimental results show that our method outperforms the existing one-branch and multi-branch networks and achieves state-of-the-art performances on three ReID benchmarks and two large-scale occluded ReID benchmarks. Additionally, the ablation study also validates our scheme can significantly improve performance in various scenes.


\bibliographystyle{cas-model2-names}

\bibliography{cas-refs}





\end{document}